\documentclass{article}

\PassOptionsToPackage{numbers}{natbib}
\usepackage[preprint]{neurips_2021}

\usepackage[utf8]{inputenc} % allow utf-8 input
\usepackage[T1]{fontenc}    % use 8-bit T1 fonts
\usepackage{hyperref}       % hyperlinks
\usepackage{url}            % simple URL typesetting
\usepackage{booktabs}       % professional-quality tables
\usepackage{amsfonts}       % blackboard math symbols
\usepackage{nicefrac}       % compact symbols for 1/2, etc.
\usepackage{microtype}      % microtypography
\usepackage{xcolor}         % colors
\usepackage{multirow}

\usepackage{amsmath}
\usepackage{graphicx}

\newcommand{\secc}[1]{Section~\ref{sec:#1}}
\newcommand{\fig}[1]{Figure~\ref{fig:#1}}
\newcommand{\tab}[1]{Table~\ref{tab:#1}}

\renewcommand{\vec}{\mathbf}
\newcommand{\ch}{\mathcal}

\title{Staircase Attention for Recurrent Processing of Sequences}

\author{Da Ju \quad Stephen Roller \quad Sainbayar Sukhbaatar \quad Jason Weston \\
\\
 Facebook AI Research
}

\begin{document}

\maketitle

\begin{abstract}
Attention mechanisms have become a standard tool for sequence modeling tasks, in particular by stacking self-attention layers over the entire input sequence as in the Transformer architecture. In this work we introduce a novel attention procedure called {\em staircase attention} that, unlike self-attention,  operates across the sequence (in time) recurrently processing the input by adding another step of processing. A step in the staircase comprises of backward tokens (encoding the sequence so far seen) and forward tokens (ingesting a new part of the sequence), or an extreme {\em Ladder}~version with a forward step of zero that simply repeats the Transformer on each step of the ladder, sharing the weights. 
We thus describe a family of such models that can trade off performance and compute, by either increasing the amount of recurrence through time, the amount of sequential processing via recurrence in depth, or both.
Staircase attention is shown to be able to solve tasks that involve tracking that conventional Transformers cannot, due to this recurrence. Further, it is shown to provide improved modeling power for the same size model (number of parameters) compared to self-attentive Transformers on large language modeling and dialogue tasks, yielding significant perplexity gains. 
 
\end{abstract}

\section{Introduction}

Early breakthrough work in neural language modeling considered a fixed context size of tokens that are embedded with a lookup table, followed by nonlinearities and a final softmax to produce a probability distribution for the next output token in a sequence 
\citep{bengio2003neural}. Such models were replaced, pre-Transformer, with recurrent models such as RNNs and LSTMs \citep{Elman1990FindingSI,hochreiter1997long,mikolov2010recurrent}  that were able to consider arbitrary context length via the ability to store state in their memory using recurrent steps through the data, in contrast to the fixed length constraint of earlier models. Moreover, the repeated application of the recurrent network across the sequence also made the models considerably deeper: a given representation is a function of a large number of nonlinearities due to previous state. This allows such models to track state, store long-term memories, and potentially solve highly nonlinear sequential tasks.
Today, with the advent of attention-based models \citep{bahdanau2014neural} and in particular Transformers \citep{vaswani2017attention}, fixed length inputs that eschew recurrence are back as the norm, thanks mainly due to deep stacks of nonlinearities on those fixed inputs that are also well suited to modern hardware, leading the authors of  \citep{vaswani2017attention} to claim that non-recurrent attention is ``all you need.'' However, some of the advantages just mentioned of earlier models -- tracking state, and solving highly nonlinear sequential tasks -- have to some degree been lost.

In this work we introduce a family of recurrent models that utilize a novel attention procedure called {\em staircase attention}.
We show that our new family of models, which utilize both sequence aligned recurrence (in time) and 
recurrence in depth can bring advantages to modern models, in particular in terms of lower  language modeling perplexities given the same number of parameters, and for solving nonlinear state-tracking tasks.
Staircase attention, like self-attention, processes tokens in parallel for speed, but 
unlike self-attention,  operates across the sequence (in time) 
recurrently processing the input by adding another step of processing. A step (processing block) in the staircase comprises of backward tokens (encoding the sequence so far seen) and forward tokens (ingesting a new part of the sequence). Thus, on each time step the block moves forward in time, retaining a memory comprised of multiple vectors stored in the backward tokens (the recurrent tokens). The blocks utilize the same model weights for each step, hence giving a recurrence in depth. An extreme  ``Ladder'' version of the blocks with a forward step of zero that simply repeats the Transformer on each step of the Ladder, sharing the weights, does not have time recurrence, but still retains depth recurrence. Compared to Transformers, utilizing staircase attention can retain a recurrent memory in time, and repeated application of the recurrent network over the sequence also makes the model considerably deeper for the same number of parameters (but not necessarily the same amount of compute).

We show on two tasks requiring state-tracking that staircase attention models can perform successfully where Transformers fail. We then show on two language modeling and a dialogue modeling task for the same number of parameters, significantly lower perplexities can be obtained compared to standard Transformers for certain models in our family. We thus analyze our family of models and show that one can control the amount of recurrence and depth which can trade off compute for performance, which has practical implications depending on available memory and compute architecture constraints. Notably, Ladder models are surprisingly effective and simple to implement on standard modeling tasks which do not appear to require much recurrence in time, but fail on some state-tracking tasks. Our Staircase models perform well on both state-tracking tasks and language modeling tasks, providing good performance across the board.

\section{Related Work}

Since attention was introduced \citep{bahdanau2014neural}, non-recurrent models have superseded in popularity recurrent models like RNNs and LSTMs \cite{Elman1990FindingSI,hochreiter1997long}, which were for a time dominant in NLP applications, particularly when involving sequence generation. The first models to use
stacked layers of attention over input word embeddings and position encodings,
as a replacement to recurrence in time,
were end-to-end memory networks \citep{sukhbaatar2015end}. Those models were shown on the task of language modeling to perform well compared to LSTMs, but in experiments still shared some weights across depth, which we refer to as  {\em recurrence in depth} (also referred to as a ``recurrent attention mechanism'' in  \cite{vaswani2017attention}). Transformers \citep{vaswani2017attention} removed the constraint of sharing any weights among layers at all, and showed this improves performance (at the cost of using more parameters). Transformers additionally contributed other notable improvements such as multi-head, self-attention and residual blocks. Such models do not have any recurrence at all, and are the current state-of-the-art architecture choice for many tasks.

Since then, several variants of Transformers have arisen that attempt to incorporate recurrence again by sharing some weights. The Universal Transformer proposes an extreme variant of applying the same layer (with shared weights) repeatedly \citep{dehghani2018universal}. Similarly, ALBERT \citep{lan2019albert} also shares the same weights across all layers for a pretraining/finetuning setting where the BERT model \citep{devlin2018bert} is effectively compressed; they also consider sharing only the self-attention or only the feed-forward weights.
We note also that several works, while not sharing parameters of layers, have studied the ordering of the sublayers of Transformers, in particular Sandwich \citep{press2019improving} and Macaron \citep{lu2019understanding} Transformers.

Some works have also attempted to incorporate sequence-aligned recurrence to Transformers.
\cite{chen2018best} incorporate LSTM layers into the Transformer, and \cite{hao2019modeling} blend a non-recurrent and recurrent model (e.g., an RNN) together with a gating function. 
Transformer-XL \citep{dai2019transformer} employs a segment-level recurrence mechanism to effectively cache and speed up computations in long-context sequence tasks. We note that a number of recent architectures have also focused on allowing long-context in Transformers, although typically without employing recurrence \citep{child2019generating,kitaev2019reformer,beltagy2020longformer}. 
Finally, the Feedback Transformer \citep{fan2020addressing}, perhaps the most similar work to ours, incorporates step-wise recurrence in the Transformer, with a step size of one and a fixed cached memory in the past. It achieves good results but has relatively high computational cost due to its architecture not fully exploiting parallelism. 

In this work, we compare architectures with the number of model parameters fixed, and explore increasing recurrence and/or compute given that fixed budget. An orthogonal topic of study is to fix the compute budget instead, but do not fix the amount of parameters, e.g. research into large, sparse (typically non-recurrent) models that may require to be spread over a cluster \cite{fedus2021switch,lewis2021base}. We focus on the former here, but learnings from each direction should be complementary.

\section{Method}

\subsection{Background}
In this paper, we consider decoder-only Transformers~\cite{al2018character,dai2019transformer} that are applied to sequential tasks like language modeling. In this setting, a Transformer model takes as  input a sequence $\{x_1, x_2, \dots , x_T \}$ of $T$ tokens and outputs a sequence of the same size
\begin{equation}
y_1, y_2, \ldots, y_T = \textsc{Transformer}(x_1, x_2, \ldots , x_T) .
\end{equation}
If we separate out the input embedding $\vec{h}_t=f_\text{in}(x_t)$ and the final output module $y_t=f_\text{out}(\vec{\bar{h}}_t)$, we are left with the Transformer core
\begin{equation}
\vec{\bar{h}}_1, \vec{\bar{h}}_2, \ldots , \vec{\bar{h}}_T = \textsc{TransCore}( \vec{h}_1, \vec{h}_2, \ldots , \vec{h}_T), 
\end{equation}
which consists of $L$ layers that compute final hidden states for each token.
Each layer is composed of self-attention and feedforward sublayers. 
In the self-attention sublayer, causal masking is applied to prevent tokens from attending to any future token, and we use relative position embeddings~\cite{shaw2018self}.
See \cite{vaswani2017attention} for more details about the sublayer architecture of Transformers.

\begin{figure}
    \begin{tabular}{c}
         \includegraphics[width=0.99\linewidth]{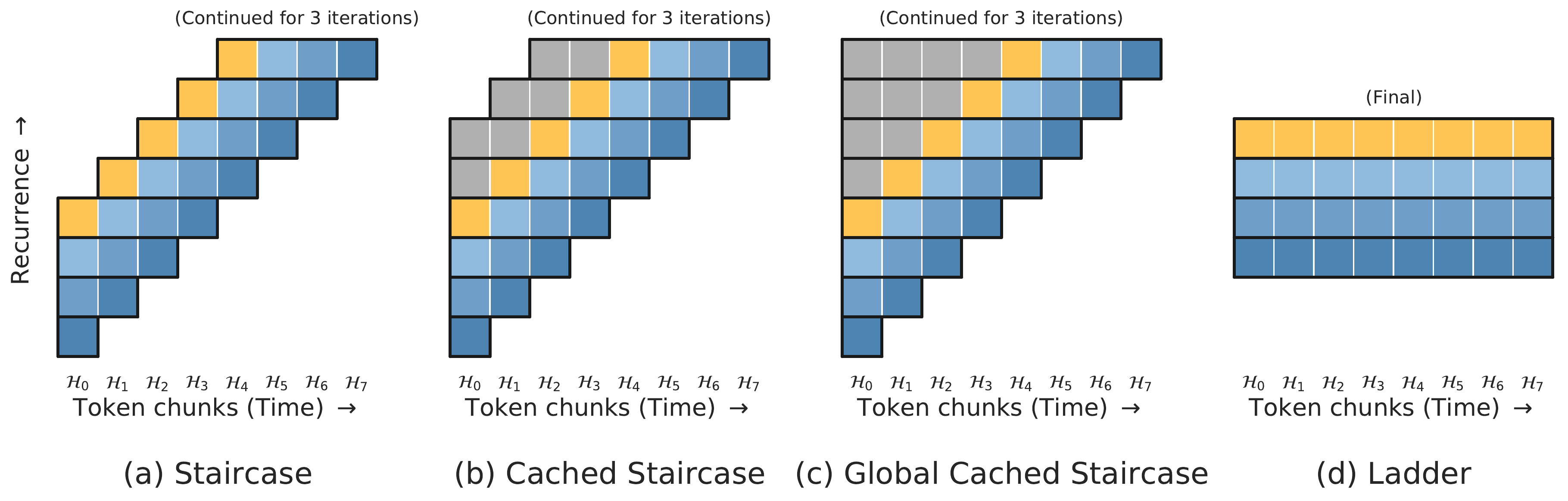}\\
         \includegraphics[width=0.6\linewidth]{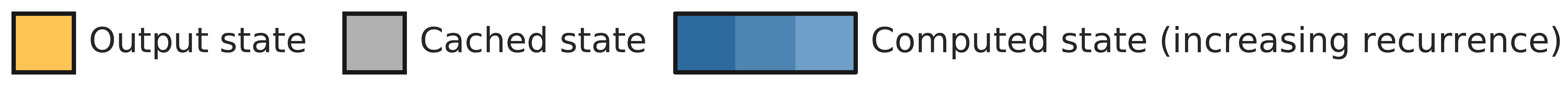}\\
    \end{tabular}
    \caption{
        \textbf{Representations of our proposed models.} Each outlined row is a parallel computation, and rows are computed recurrently from bottom to top.
        (a) In the {\em Staircase} model, on every time step the model sees one new input chunk and performs recurrent computation on a fixed number of previous chunks.
        (b) In the {\em Cached Staircase}, the final output state is cached and included within the attention span of later chunks after some fixed amount of recurrent processing.
        (c) In the {\em Global Cached Staircase}, all previous chunks are cached and attended in the final chunk.
        (d) In the {\em Ladder} model, the full sequence is fed in without chunking, and the model is repeated a fixed number of times.
    }
    \label{fig:proposedmethods}
\end{figure}

\subsection{Staircase Attention Family of Models}
We now describe our family of models that utilize staircase attention. 
Their graphical representation may be found in Figure~\ref{fig:proposedmethods}.
We start with Staircase models that are recurrent in both time and depth.

\subsubsection{Staircase Model}
Unlike Transformers, a Staircase model ingests input tokens in smaller chunks, as shown Figure~\ref{fig:proposedmethods}a.
Inside a Staircase model lies a Transformer core that processes each input token in $N$ recurrent steps.
With each recurrent step, a Staircase model moves $C$ tokens forward in time, which we call the \emph{forward} step size. In addition to those $C$ forward tokens, the model simultaneously also processes $NC-C$ tokens that come from the previous step, which we call \emph{backward} tokens. We refer to the total number of tokens $NC$ that are being processed in parallel as the \emph{step size}.

Let us denote a chunk of C input tokens as $\ch{H}_{i}^0 = \{ \vec{h}_{iC+1}, \ldots, \vec{h}_{iC+C} \}$.
Here $\vec{h}_t$ is the embedding vector of the token $x_t$. 
At each step, the Transformer core processes $N$ chunks in parallel
\begin{equation}\label{eq:stair}
\ch{H}_{i-N+1}^{N}, \ldots, \ch{H}_{i-1}^2, \ch{H}_{i}^1 = \textsc{TransCore}(\ch{H}_{i-N+1}^{N-1}, \ldots, \ch{H}_{i-1}^1, \ch{H}_{i}^0) .
\end{equation}
Among the input chunks, only $\ch{H}_{i}^0$ contains new token embeddings while the remaining $N-1$ chunks come from the previous recurrent step.
The superscript $n$ of $\ch{H}_i^n$ denotes the number of computational passes through the Transformer core. 
After $N$ passes through the core module, the output for a particular token $x_t$ is computed with
\[
y_t = f_\text{out}(\vec{\bar{h}}_t) \quad \text{for all} \; \vec{\bar{h}}_t \in  \ch{H}_i^{N} .
\]

As you can see, an input token gets processed by the same core module $N$ times. This makes it possible to control the amount of computation by varying the number of recurrent steps $N$ without changing the number of parameters of the model. 

Feeding states computed by the previous step into the next step 
computation makes Staircase models recurrent in time like RNNs  because each recurrent step moves forward $C$ tokens.
There are two advantages to this type of recurrence.
First, the number of non-linear computations between an input token $x_t$ and output token $y_{t+\tau}$ linearly increases with their distance $\tau$.
In contrast, Transformers are strictly a feedforward model that has a fixed number of computation steps.
The second advantage is that information can pass to future steps without any limits, whereas standard Transformers are limited by their token truncation length.
These two advantages make recurrent models capable of maintaining an internal state, but more importantly of  updating this state continuously through time. However, unlike RNNs, Staircase models take advantage of the attention mechanism in the same way as Transformers, and store state across multiple vectors: the $NC - C$ backward tokens. Like Transformers, they thus take advantage of parallelism.

\subsubsection{Cached Staircase Model}
\label{sec:cached}
In Staircase models, the self-attention sublayer processes $NC$ tokens at a time.
This means how far a token can directly attend to is limited by this context size $NC$.
However, the hyperparameter $N$ also controls the number of recurrent computations, and one might want to decouple these two factors to control context size versus recurrence.

Here we propose a simple solution for increasing the context size while keeping the recurrent computation constant, shown graphically in Figure~\ref{fig:proposedmethods}b.
We do this by introducing a new hyperparameter $M < N$ and put hidden states in a \emph{cache} after $M$ recurrent steps. Once a hidden state is in the cache, it stays the same, requiring no additional computation
\[
\ch{H}_i^{n} = \ch{H}_i^M \quad \text{for} \; n > M .
\]
This means the number of recurrent computations on a particular input is limited to $M$.
However, hidden states stay in the cache for the remaining $N-M$ steps so other tokens still can attend to them. 
This is achieved by including cached hidden states only when computing keys and values in the self-attention sublayer of the Transformer core.
As a result, the self-attention sublayer will have $NC$ keys and values, but only $MC$ queries, reducing its computational complexity from $\mathcal{O}(N^2C^2)$ to $\mathcal{O}(NMC^2)$.
As cached hidden states are excluded from the feedforward sublayer altogether, the computational complexity there changes from $\mathcal{O}(NC)$ to $\mathcal{O}(MC)$.
Thus, the context size $NC$ can be increased by picking a larger $N$, but the amount of computation can be reduced by choosing a smaller $M$.
For example, for $M=1$, we can see that the reduction in computation is $N$ fold.

\subsubsection{Global Cached Staircase Model}

For sequence lengths that are not excessively long, it may be desirable at any stage of computation to always have access to all the tokens from the past, whereas the models discussed so far have the limit of $NC$ tokens, see Figures~\ref{fig:proposedmethods}a and \ref{fig:proposedmethods}b.
We can extend the Cached Staircase model to look back across all tokens, called the Global Cached Staircase.
This is achieved by increasing $N$ by one with each step, so {\em all} prior representations $\ch{H}_i^M$ are in the cache and available during later computations, shown in Figure~\ref{fig:proposedmethods}c. We still employ the cache hyperparameter $M$ as before to control the amount of recurrence and computation necessary during the steps of processing.

\subsubsection{Ladder Model}
Here we introduce a special variant of the Staircase model, shown in Figure~\ref{fig:proposedmethods}d.
The forward step size $C$ is a variable that we can freely adjust. 
Instead of taking the same small forward steps each time, we consider an extreme case where the model ingests all tokens at once and then sets the forward step $C$ to 0 for the remaining $N-1$ steps. 
With this choice, the model is recurrent  only in its depth and lacks recurrence in time, so we call it the \emph{Ladder} model.
This method can be viewed as applying the Transformer core multiple times on the given sequence of input tokens, on each application taking the output hidden state $\{ \ch{H}_{1}^{n}, \ch{H}_{2}^{n}, \ldots, \ch{H}_{T}^n \}$  of the $n^{th}$ application as input for the next ${n+1}^{th}$ application (here, viewing each individual token as belonging to its own separate chunk). In this way, it may also be seen as a generalization of previous methods with shared weights \cite{dehghani2018universal,lan2019albert}.

We use the same techniques as Transformer-XL~\cite{dai2019transformer} for processing very long or unbounded sequences.
First, each token will attend to a fixed number of previous tokens $S$ rather than the whole sequence. 
This reduces the computational complexity of the self-attention from $\mathcal{O}(T^2)$ to $\mathcal{O}(TS)$ assuming $S \ll T$.
Next, we split the input sequence into smaller manageable segments and let the model process them sequentially.
To avoid the segment boundaries from obstructing attention across segments, the hidden states computed in the previous segments are kept in cache (different from the cache in \secc{cached}).
Then, this cache is made available in the self-attention sublayer for subsequent segments so a token can attend to a token in the previous segment.
See~\cite{dai2019transformer} for more details about this caching mechanism.

\subsection{Relation to Existing Models}

\paragraph{Transformer}  The standard Transformer corresponds to a Staircase model with a large chunk size and no recurrence, or equivalently a Ladder model with only $N=1$ steps (i.e, no recurrence). While it is efficient at processing tokens in parallel, it has no ability of retaining and recomputing state across sequences, other than by fitting those tokens into the current processing block.

\vspace{-2.5mm} 
\paragraph{Feedback Transformer}  The Feedback Transformer \citep{fan2020addressing} is equivalent to a Cached Staircase model with a chunk size of $C=1$ (i.e., a forward step of a single token), and caching after $M=1$ step, i.e. all tokens in the past are part of a fixed cached memory. In contrast, larger chunk sizes allow general Staircase models to exploit parallelism and be more efficient than the Feedback Transformer, while increasing $M$ can give more expressive power. We compare  to this model in our experiments.

\vspace{-2.5mm} 
\paragraph{Recurrent Neural Networks} RNNs \cite{Elman1990FindingSI} that store recurrent state in a single vector and ingest tokens one at a time can be compared to a Staircase model with a single backward token and a single forward token, i.e. a chunk size of $C=1$ and $N=2$. Staircase models exploit parallelism similar to Transformers while maintaining several chunks of recurrent (per token) features to more expressively track state than conventional RNNs.

\vspace{-2.5mm} 
\paragraph{Memory Networks} MemNets as implemented in \citep{sukhbaatar2015end} employ recurrence in the stacked layers of attention and computation for the current token, but only compute input embeddings $\vec{h}_t^0=f_\text{in}(x_t)$ for previous tokens, and can thus be seen as a kind of simplified Ladder model, or equivalently a Global Cached Staircase with a chunk size of $C=1$ and caching at all previous steps, $M=0$.

\vspace{-2.5mm} 
\paragraph{Transformer-XL} Transformer-XL, like Cached Staircase, also has a caching mechanism which eases computation when dealing with  earlier chunks of tokens. The difference is that Staircase models take the last state of earlier chunks and process that state further in a recurrent way; Transformer-XL on the other hand extends each layer of the Transformer's attention mechanism to using old cached states at each layer, i.e. does not build further computations on top of the old state. 
We use this as a baseline in our experiments.

\vspace{-2.5mm} 
\paragraph{Universal Transformer} Universal Transformers \citep{dehghani2018universal} propose to tie all the  layer weights in a Transformer, which can be seen as a Ladder model with a Transformer core of a single layer.
We also compare to this model.

\section{Experiments}

\begin{table}[t]
    \centering
    \small
    \caption{{\bf Results summary across all our tasks.} We compare five architectures where we fix the number of learnable parameters to be the same for all models on the same task.  
    }
    \begin{tabular}{lccccc}
    \toprule
         & Random Walk & Algorithm & Enwik8 & Reddit & BASE Data \\
        Model & (error \%) & (error \%) & (test bpc) & (test ppl)  & (valid ppl) \\
        \midrule
        Transformer-XL~\citep{dai2019transformer} & 84.1 & 48.7 & 1.15 & 26.2 & 28.0 \\
        Feedback Transformer~\cite{fan2020addressing} & \bf 0.1 & \bf 0.2 & 1.12 & 25.5 & 26.6 \\
        \midrule
        {\em Our models} \\
        Staircase & \bf 0.1 & \bf 0.2 & 1.14 & 22.6 &  23.0 \\
        Cached Staircase & \bf 0.1 &  1.2 & 1.13 & 26.1 & 27.8\\ 
        Ladder & \bf 0.1 & 45.1 & \bf 1.11 & \bf 22.5 & \bf 22.9\\ 
        \bottomrule
    \end{tabular}
    \label{tab:main-results}
\end{table}

\begin{figure}[t]
    \centering
    \includegraphics[width=0.45\linewidth]{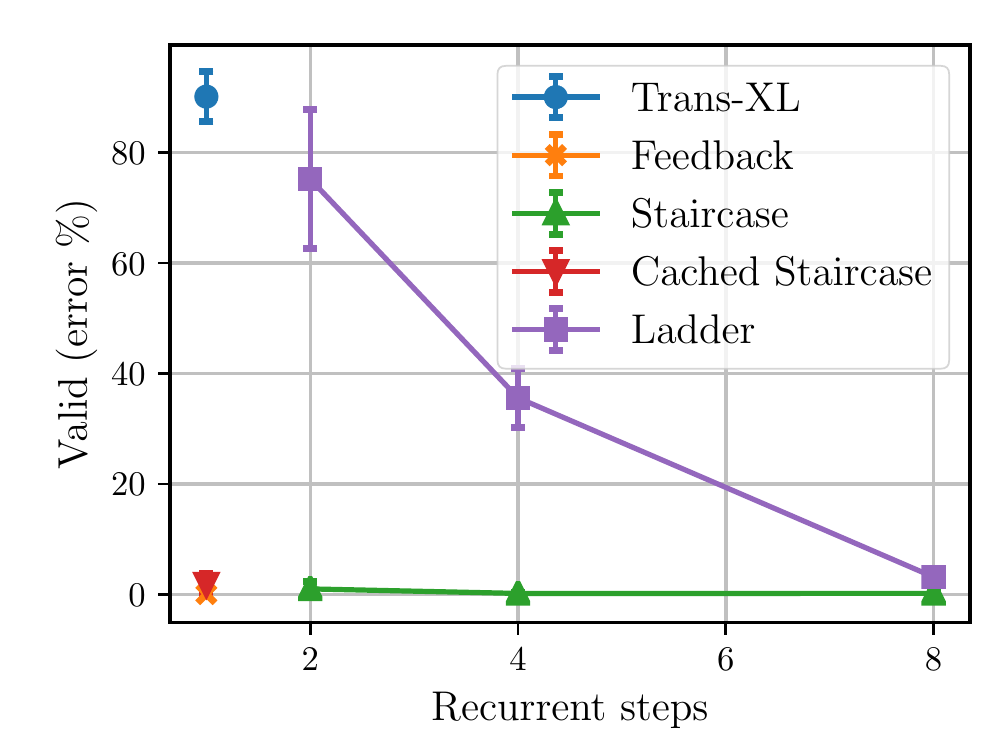}
    \hspace{5mm}
    \includegraphics[width=0.45\linewidth]{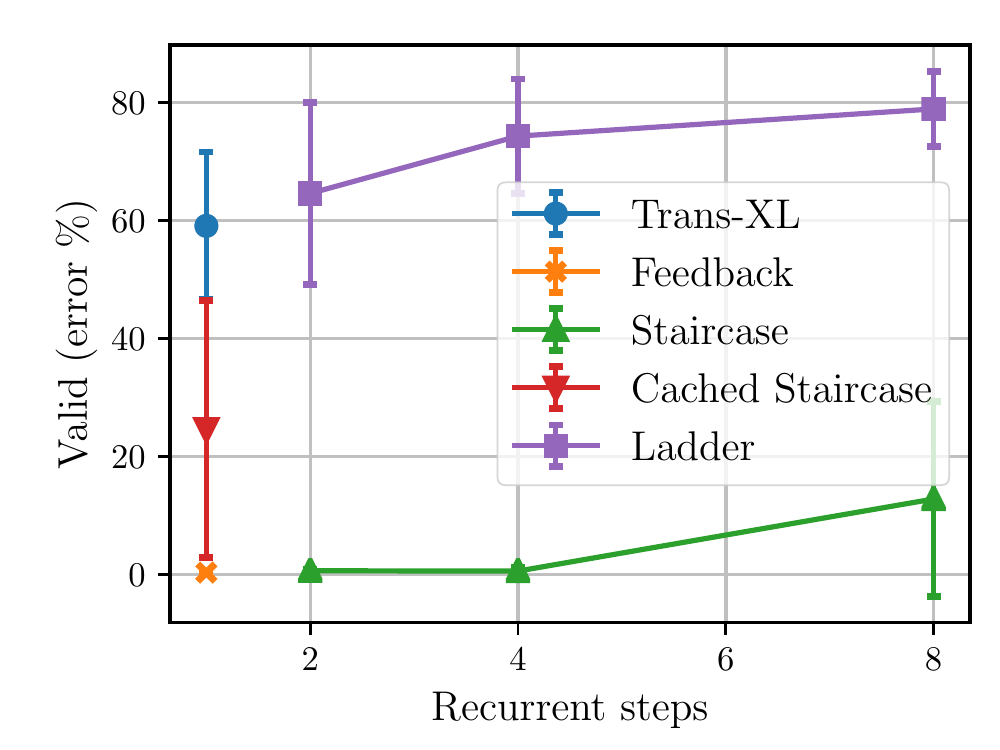}
    \caption{The performance of various models on \textbf{Random Walk (left)} and \textbf{Algorithm (right)} tasks as we increase the number of recurrent steps. Note that the number of parameters is the same for all the models.
    We run each experiment multiple times and plot their mean and standard deviations.}
    \label{fig:toy}
\end{figure}

We use two types of tasks to test our family of models and compare its variations, along with Transformer-XL~\cite{dai2019transformer} and Feedback Transformer~\cite{fan2020addressing} baselines.
First, we have two artificial state tracking tasks specifically designed to test the model's ability to keep track of evolving changes. Next, we use real-world language modeling tasks.
See the supplementary material for further details of our experimental setup for training, including all hyperparameter choices.

\subsection{Tasks}
\label{sec:tasks}

\paragraph{Random Walk}
At each time step, an agent in a small grid takes a random action that turns the agent, or moves it forward. A model has to predict the agent's location given these actions.
This seemingly simple task requires recurrency and has been shown to  make feedforward models like Transformers fail.
We borrow this task from~\cite{fan2020addressing}, but increase the length from 100 to 400 to make it more challenging. 
See~\cite{fan2020addressing} for more details about this task.

\vspace{-2.5mm} 
\paragraph{Algorithm}
This task consists of simple algorithmic operations that need to be executed sequentially.
Some operations depend on the current variable values, which makes it necessary to keep track of variable values and update them if necessary.
Thus, it also requires recurrency, and like the Random Walk task has been shown to make Transformers fail.
However, it is more complex and also cannot be solved with LSTMs~\cite{fan2020addressing}.
We use the 3 variable version of the task from ~\cite{fan2020addressing}.

\vspace{-2.5mm} 
\paragraph{Enwik8}
Enwik8 is a character-level language modeling task~\cite{mahoney2011large} that consists of 100M tokens taken from Wikipedia articles. The challenging part of this data is that it requires long-term reasoning~\cite{sukhbaatar2019adaptive} because tokens are characters instead of words.

\vspace{-2.5mm} 
\paragraph{Pushshift.io Reddit}
We use a variant of Reddit discussions, which has also been used in several existing studies, see e.g. \cite{reddit_use, mazare2018trainingmillions,keskar2019ctrl,shuster2019dialogue}.
Following \cite{humeau2019polyencoder}, we use a previously existing Reddit dataset extracted and obtained by a third party and made available on pushshift.io \cite{baumgartner2020pushshift},
training to generate a comment conditioned on the full thread leading up to the comment, spanning 1.5B training examples from Reddit obtained from Pushshift through July 2019.  See \cite{roller2020recipes} for more details. We concatenate the dataset to view it as a language modeling task.

\vspace{-2.5mm} 
\paragraph{BASE Data}
 We use the language modeling dataset from \cite{lewis2021base}, which consists of approximately 100B tokens, combining the corpora used in \cite{liu2019roberta} that consists of Wikipedia, BookCorpus, CC-News, OpenWebTex and Stories, along
 with the English subset of the CC100 corpus \cite{conneau2019unsupervised}. 

\subsection{Results}

\begin{table}[t]
  \caption{{\bf Detailed performance on Pushshift.io Reddit.} We compare our family models with varying recurrent steps to baselines with the same number of parameters. Additionally, a twice as deep (2x) baseline is included, and Universal Transformers with the same layer size.}
  \label{tab:reddit-performance}
  \small
  \centering
  \begin{tabular}{lrcccccr}
    \toprule
    Model & Num. of & Recurrent & Step & Forward & Valid. & Test & Train batch \\
    & params & steps & size & size & (ppl) & (ppl) & time (ms) \\
    \midrule
    Transformer-XL~\cite{dai2019transformer} & 117M & - & - & - & 26.5 & 26.2 & 178\\
    Transformer-XL (2x) &  218M & - & - & - & 23.7 & 23.4 & 359 \\
    Feedback Transformer~\cite{fan2020addressing} & 102M & 1 & 512 & 1 & 25.8 & 25.5 & 3260 \\
    Cached Staircase & 117M  & 1 & 256 & 128  & 26.4 & 26.1 & 246\\
    \midrule
    Staircase & 117M & 2 & 256 & 128 & 25.0 & 24.8 & 297\\
    Ladder& 117M & 2 & - & - & 25.0 & 24.7 & 328 \\
    Universal Transformer& 29M & 2 & - & - & 39.9 & 39.5 & 51 \\
    \midrule
    Staircase & 117M & 4 & 256 & 64 & 23.7 & 23.4 & 580 \\
    Ladder & 117M & 4 & - & - & 23.6 & 23.3  & 627 \\
    Universal Transformer& 29M  & 4 & - & - & 34.9 & 34.5 & 88\\
    \midrule
    Staircase&  117M & 8 & 256 & 32 & 22.9 & 22.6 & 1147\\
    Ladder & 117M & 8 & - & - & 22.8 & 22.5 & 1232 \\
    Universal Transformer& 29M  & 8 & - & - & 32.3 & 32.0 & 163 \\
    \bottomrule
  \end{tabular}
\end{table}

\begin{figure}[t]
    \centering
    \includegraphics[width=0.45\linewidth]{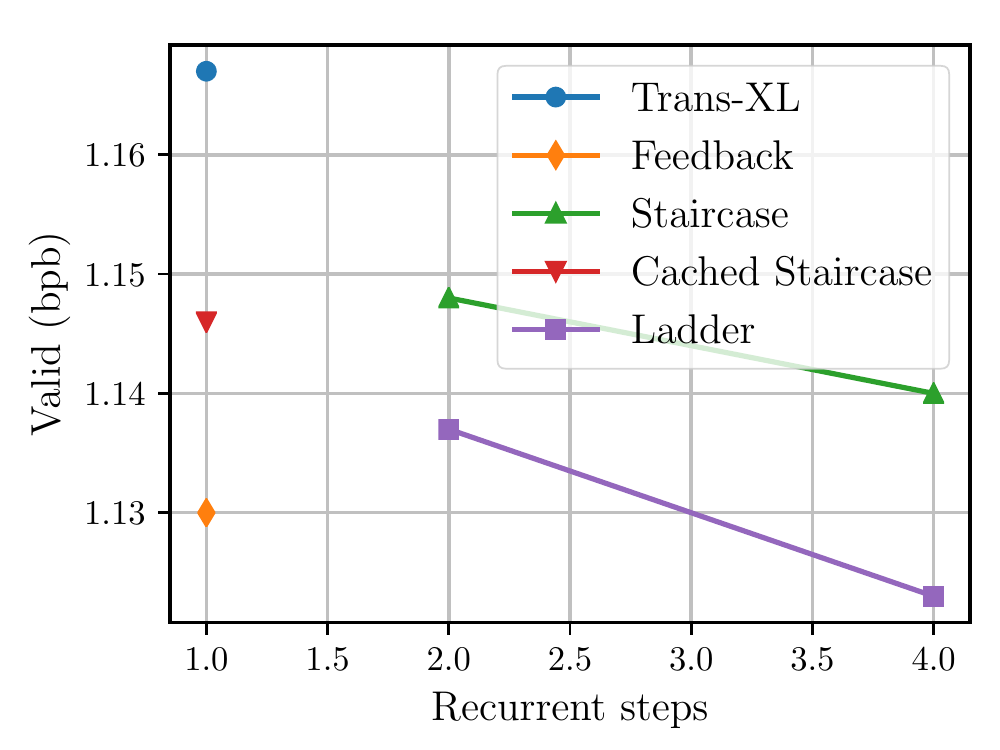}
    \includegraphics[width=0.45\linewidth]{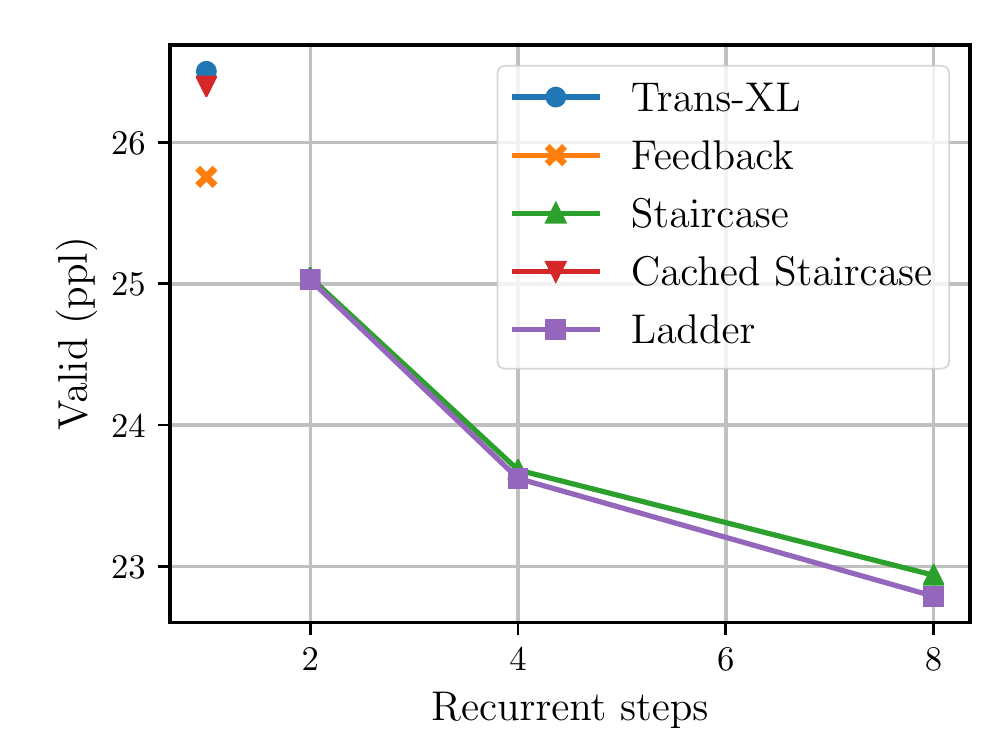}
    \caption{The performance on the language modeling tasks \textbf{Enwik8 (left)} and \textbf{Pushshift.io Reddit (right)} as we increase the recurrent steps. All models have the same number of parameters. }
    \label{fig:lm-tasks}
\end{figure}

Our results on all of the tasks are summarized in \tab{main-results}. For each task, all the models have the same number of parameters and use the same Transformer core architecture implementation.
For Random Walk and Algorithm tasks, we trained each model with multiple seeds and chose the best seed as measured by their validation performance.
The specific configuration of each model can be found in Table \ref{app:best_params} in the supplementary material.
We see clear wins from our Staircase attention family of models on all the tasks, and the following subsections will analyze these results in detail.

\subsubsection{Staircase models have strong performance on state tracking tasks}
The Random Walk and Algorithm tasks are specifically designed to test a model's capability of tracking states: to store given information internally and update it with new information coming at each time step.
In \tab{main-results} we report results from running multiple training seeds, and selecting the one with best performance on the validation set.
In \fig{toy} we show detailed results when varying the recurrent computation steps, reporting the mean and standard deviations amongst the seeds.

The powerful Transformer-XL baseline performs poorly here due to its lack of a recurrent mechanism, confirming the results from~\cite{fan2020addressing}.
The self-attention can access a hidden state far away in the past, but updating that hidden state with a new piece of information brings it up one layer higher.
Thus, in a Transformer with $L$ layers, a particular hidden state can be updated only $L$ times before it reaches the final output layer, and becomes unavailable for future computations.
This limited computation depth is a problem in the Random Walk task, for example, because the model needs to internally store the agent's location and update it with actions taken at every time step for hundreds of steps.

The Ladder model does perform better with more recurrent steps on the Random Walk task, eventually solving it with 8 steps. 
With $N$ recurrent steps, a token gets processed by $NL$ layers. This, in turn, also increases the number of updates that can be applied to a particular hidden state. However, this was not sufficient on the harder Algorithm task, where the Ladder model failed.

The Staircase model successfully solves both tasks, even with only two recurrent steps.
Thanks to its recurrence through time, the computation depth is only restricted by the input sequence length itself.
More concretely, each recurrent step can update the output from the previous step and feed it to the next step, making it possible to maintain and update an internal state without limit.

The Cached Staircase model also performs reasonably well on those tasks.
While we only ran this model with $M=1$ computation step, it is still recurrent in time which is more critical for these tasks than increased computation.
The Global Cached Staircase models did not perform any better, see \autoref{tab:algorithm-performance} and \ref{tab:random_walk} in the supplementary material, so we do not consider them in further experiments.

The Feedback Transformer solves both tasks, which is not surprising as it is a particular case of a Cached Staircase model with a forward step $C=1$. However, such fine-grained steps make it slow to train in practice because of the reduced parallelism, as we will see in the analysis in the next section.

\begin{table}[t]
  \caption{Staircase model performance on Pushshift.io Reddit with different forward sizes.}
  \label{tab:reddit-step-size}
  \small
  \centering
  \begin{tabular}{lccccc}
    \toprule
    Model & Recurrent & Step & Forward & Valid. & Train batch \\
    & steps  & size & size & (ppl) & time (ms) \\
    \midrule
    Cached Staircase & 1 & 288 & 32  & 26.4 & 489 \\
    Cached Staircase & 1 & 320 & 64  & 26.4 & 318 \\
    Cached Staircase & 1 & 384 & 128 & 26.4 & 246\\
    \midrule
    Staircase & 2 & 128 & 64 & 25.2 & 310 \\
    Staircase & 2 & 256 & 128 & 25.0 & 297 \\
    \midrule
    Staircase & 4 & 128 & 32 & 23.8 & 605 \\
    Staircase & 4 & 256 & 64 & 23.7 & 580 \\
    \bottomrule
  \end{tabular}
\end{table}
\begin{table}
  \caption{Comparison of Ladder models and layer composition patterns on Pushshift.io Reddit.}
  \label{tab:reddit-layer-recurrence}
  \small
  \centering
  \begin{tabular}{lcccc}
    \toprule
    Layer & Num. of & Recurrent & Num. & Valid.\\
    pattern & params & steps & layers & (ppl) \\
    \midrule
    {\small{AB}}   & 117M & - &  8 & 26.5 \\
    \midrule
    {\small{ABCD}} & 218M & - & 16 & 23.7 \\
    {\small{ABAB}} &117M & 2&  16 &25.0 \\
    {\small{AABB}} &117M & 2&  16 & 24.8 \\
    \midrule
    {\small{ABABABAB}} &117M & 4&  32 & 23.6  \\
    {\small{AAAABBBB}} &117M & 4&  32 & 23.8 \\
    \midrule
   {\small{ABABABABABABABAB}} &117M & 8&  64 & 22.8 \\
    \bottomrule
  \end{tabular}
\end{table}

\subsubsection{Staircase attention models outperform Transformers for the same number of parameters on language modeling tasks}

\tab{main-results} shows results on the three language modeling tasks, Enwik8, Pushshift.io Reddit and BASE Data. We show performance versus recurrence plots for the
first two tasks in particular in \fig{lm-tasks}.
We also show more detailed performance numbers on the Pushshift.io Reddit task in \tab{reddit-performance}.
In all three tasks, one general trend is that more recurrent steps improve the performance significantly.
On the Pushshift.io Reddit task, we see a $\sim$4 perplexity point improvement over the Transformer-XL baseline without adding any new parameters when using 8 recurrent steps, and a $\sim$5 point improvement on BASE Data.
Making a twice as deep Transformer-XL (marked with ``2x'' in \tab{reddit-performance}) improves the baseline at the expense of having twice as many parameters than the Staircase and Ladder models, but is still $\sim$1 perplexity point worse, showing the power of our recurrent models.
Our family of models provides a new way of improving model performance without increasing the number of parameters that is generally applicable.

On Enwik8, the Ladder models outperformed the Staircase models for matching recurrent steps, whereas on Pushshift.io Reddit and BASE Data they are about even. This could be due to the long context requirement of the character-level data of Enwik8.
The Staircase model tries to compress past context into a fixed number of hidden states, equal to $NC-C$ backward tokens to be precise.

\tab{reddit-performance} also shows the time it takes for training on a single batch for each model. Models with more recurrent steps take longer to train as they have to perform more computations per token, but are still tractable and much faster than the Feedback Transformer. The Feedback Transformer does not perform more computations, but it is slow because it processes one token at a time instead of $C$ tokens, and also generally performs worse in our language model experiments.
The Staircase model is slightly faster than the Ladder models because it generally has a smaller context size.

\subsubsection{Staircase model’s forward size and step size control its performance}

The forward step chunk size $C$ and overall staircase step size $NC$ are hyperparameters in Staircase models, where the effective number of recurrent steps is determined by those choices of $N$ in the Staircase model, or truncated to only $M$ steps due to caching in the Cached Staircase model.
In \tab{reddit-step-size}, we compare different values of
step size and forward size on the 
Pushshift.io Reddit task for those models 
with differing numbers of recurrent steps.
We see that, in general, 
the models are robust to different choices of those values.
Larger forward step sizes are preferable in terms of computational efficiency because they allow more parallelism, but if they are too large some performance in terms of perplexity is lost. We see  evidence of this in
\autoref{tab:algorithm-performance} and \autoref{tab:random_walk} in the supplementary material where the Cached Staircase model performs poorly as its forward size $C$ increases.

\subsubsection{Ladder model recurrence analysis}
In this section we analyze the Ladder model by comparing it against some alternatives. 
Let us consider a Transformer core with layers A and B as an illustrative example.
First, a Ladder model with 2 recurrent steps, i.e. the pattern ABAB, can be compared directly to a twice as big Transformer-XL baseline, i.e with a pattern ABCD. Both models in that case would be the same size and number of layers except the Ladder model has tied weights (and hence the baseline has twice as many parameters), and both would have the same test speed. As shown in \autoref{tab:reddit-layer-recurrence}, going from the baseline (AB pattern, with 8 layers) to the 2x baseline (ABCD pattern, with 16 layers) gives close to a 3 perplexity point improvement on Pushshift.io Reddit due to increasing the number of parameters, whereas a Ladder model with 2 recurrent steps (ABAB) is slightly behind but still improves over the 1x baseline by 1.5 perplexity points. However, with no extra parameters it matches the 2x baseline with 4 recurrent steps (ABABABAB), and outperforms the 2x baseline with 8 recurrent steps. 

We also try out another way of composing the layers of the Ladder model, an AABB pattern instead of ABAB -- that is, to repeat each layer multiple times consecutively before going  on to the next, instead of repeating the whole Transformer core. While there are slight differences, overall there was no clear winner. While they learn quite different functions (e.g if you take a trained model and place its learned weights in the other configuration the testing results will be poor) it seems learning can adapt well to either design. We note that the Universal Transformer (that uses the pattern AAAA\dots) performs poorly, at least when using the same layer size as these other models,  as shown in Table \ref{tab:reddit-performance}.

\section{Conclusion}
We present Staircase Attention, which re-introduces recurrence back into the family of Transformer-based models across both time and depth. We show that our Staircase models are able to solve tasks which require tracking of state that conventional Transformers cannot. Our models deliver more {\em modeling power per parameter} than conventional Transformers via their recurrence, thus also giving improved performance in language modeling tasks for the same number of parameters, which is especially important in regimes which are memory rather than compute bound. Future work should continue to investigate how recurrence can be built into sequence models, as without a memory component our systems will always be limited to only short-term reactive tasks with limited input. The approaches detailed here are one way forward and should be studied in further applications.

\bibliography{ourbib}

\begin{thebibliography}{34}
\providecommand{\natexlab}[1]{#1}
\providecommand{\url}[1]{\texttt{#1}}
\expandafter\ifx\csname urlstyle\endcsname\relax
  \providecommand{\doi}[1]{doi: #1}\else
  \providecommand{\doi}{doi: \begingroup \urlstyle{rm}\Url}\fi

\bibitem[Bengio et~al.(2003)Bengio, Ducharme, Vincent, and
  Janvin]{bengio2003neural}
Yoshua Bengio, R{\'e}jean Ducharme, Pascal Vincent, and Christian Janvin.
\newblock A neural probabilistic language model.
\newblock \emph{The journal of machine learning research}, 3:\penalty0
  1137--1155, 2003.

\bibitem[Elman(1990)]{Elman1990FindingSI}
J.~Elman.
\newblock Finding structure in time.
\newblock \emph{Cogn. Sci.}, 14:\penalty0 179--211, 1990.

\bibitem[Hochreiter and Schmidhuber(1997)]{hochreiter1997long}
Sepp Hochreiter and J{\"u}rgen Schmidhuber.
\newblock Long short-term memory.
\newblock \emph{Neural computation}, 9\penalty0 (8):\penalty0 1735--1780, 1997.

\bibitem[Mikolov et~al.(2010)Mikolov, Karafi{\'a}t, Burget, {\v{C}}ernock{\`y},
  and Khudanpur]{mikolov2010recurrent}
Tom{\'a}{\v{s}} Mikolov, Martin Karafi{\'a}t, Luk{\'a}{\v{s}} Burget, Jan
  {\v{C}}ernock{\`y}, and Sanjeev Khudanpur.
\newblock Recurrent neural network based language model.
\newblock In \emph{Eleventh annual conference of the international speech
  communication association}, 2010.

\bibitem[Bahdanau et~al.(2015)Bahdanau, Cho, and Bengio]{bahdanau2014neural}
Dzmitry Bahdanau, Kyunghyun Cho, and Yoshua Bengio.
\newblock Neural machine translation by jointly learning to align and
  translate.
\newblock In \emph{{ICLR}}, 2015.

\bibitem[Vaswani et~al.(2017)Vaswani, Shazeer, Parmar, Uszkoreit, Jones, Gomez,
  Kaiser, and Polosukhin]{vaswani2017attention}
Ashish Vaswani, Noam Shazeer, Niki Parmar, Jakob Uszkoreit, Llion Jones,
  Aidan~N Gomez, {\L}ukasz Kaiser, and Illia Polosukhin.
\newblock Attention is all you need.
\newblock In \emph{Advances in neural information processing systems}, pages
  5998--6008, 2017.

\bibitem[Sukhbaatar et~al.(2015)Sukhbaatar, Szlam, Weston, and
  Fergus]{sukhbaatar2015end}
Sainbayar Sukhbaatar, Arthur~D. Szlam, J.~Weston, and R.~Fergus.
\newblock End-to-end memory networks.
\newblock In \emph{NIPS}, 2015.

\bibitem[Dehghani et~al.(2018)Dehghani, Gouws, Vinyals, Uszkoreit, and
  Kaiser]{dehghani2018universal}
Mostafa Dehghani, Stephan Gouws, Oriol Vinyals, Jakob Uszkoreit, and {\L}ukasz
  Kaiser.
\newblock Universal transformers.
\newblock \emph{arXiv preprint arXiv:1807.03819}, 2018.

\bibitem[Lan et~al.(2019)Lan, Chen, Goodman, Gimpel, Sharma, and
  Soricut]{lan2019albert}
Zhenzhong Lan, Mingda Chen, Sebastian Goodman, Kevin Gimpel, Piyush Sharma, and
  Radu Soricut.
\newblock Albert: A lite bert for self-supervised learning of language
  representations.
\newblock \emph{arXiv preprint arXiv:1909.11942}, 2019.

\bibitem[Devlin et~al.(2018)Devlin, Chang, Lee, and Toutanova]{devlin2018bert}
Jacob Devlin, Ming-Wei Chang, Kenton Lee, and Kristina Toutanova.
\newblock Bert: Pre-training of deep bidirectional transformers for language
  understanding.
\newblock \emph{arXiv preprint arXiv:1810.04805}, 2018.

\bibitem[Press et~al.(2019)Press, Smith, and Levy]{press2019improving}
Ofir Press, Noah~A Smith, and Omer Levy.
\newblock Improving transformer models by reordering their sublayers.
\newblock \emph{arXiv preprint arXiv:1911.03864}, 2019.

\bibitem[Lu et~al.(2019)Lu, Li, He, Sun, Dong, Qin, Wang, and
  Liu]{lu2019understanding}
Yiping Lu, Zhuohan Li, Di~He, Zhiqing Sun, Bin Dong, Tao Qin, Liwei Wang, and
  Tie-Yan Liu.
\newblock Understanding and improving transformer from a multi-particle dynamic
  system point of view.
\newblock \emph{arXiv preprint arXiv:1906.02762}, 2019.

\bibitem[Chen et~al.(2018)Chen, Firat, Bapna, Johnson, Macherey, Foster, Jones,
  Parmar, Schuster, Chen, et~al.]{chen2018best}
Mia~Xu Chen, Orhan Firat, Ankur Bapna, Melvin Johnson, Wolfgang Macherey,
  George Foster, Llion Jones, Niki Parmar, Mike Schuster, Zhifeng Chen, et~al.
\newblock The best of both worlds: Combining recent advances in neural machine
  translation.
\newblock \emph{arXiv preprint arXiv:1804.09849}, 2018.

\bibitem[Hao et~al.(2019)Hao, Wang, Yang, Wang, Zhang, and Tu]{hao2019modeling}
Jie Hao, Xing Wang, Baosong Yang, Longyue Wang, Jinfeng Zhang, and Zhaopeng Tu.
\newblock Modeling recurrence for transformer.
\newblock \emph{arXiv preprint arXiv:1904.03092}, 2019.

\bibitem[Dai et~al.(2019)Dai, Yang, Yang, Carbonell, Le, and
  Salakhutdinov]{dai2019transformer}
Zihang Dai, Zhilin Yang, Yiming Yang, Jaime~G. Carbonell, Quoc~Viet Le, and
  Ruslan Salakhutdinov.
\newblock Transformer-xl: Attentive language models beyond a fixed-length
  context.
\newblock In \emph{{ACL} {(1)}}, pages 2978--2988. Association for
  Computational Linguistics, 2019.

\bibitem[Child et~al.(2019)Child, Gray, Radford, and
  Sutskever]{child2019generating}
Rewon Child, Scott Gray, Alec Radford, and Ilya Sutskever.
\newblock Generating long sequences with sparse transformers.
\newblock \emph{arXiv preprint arXiv:1904.10509}, 2019.

\bibitem[Kitaev et~al.(2019)Kitaev, Kaiser, and Levskaya]{kitaev2019reformer}
Nikita Kitaev, Lukasz Kaiser, and Anselm Levskaya.
\newblock Reformer: The efficient transformer.
\newblock In \emph{International Conference on Learning Representations}, 2019.

\bibitem[Beltagy et~al.(2020)Beltagy, Peters, and Cohan]{beltagy2020longformer}
Iz~Beltagy, Matthew~E Peters, and Arman Cohan.
\newblock Longformer: The long-document transformer.
\newblock \emph{arXiv preprint arXiv:2004.05150}, 2020.

\bibitem[Fan et~al.(2020)Fan, Lavril, Grave, Joulin, and
  Sukhbaatar]{fan2020addressing}
Angela Fan, Thibaut Lavril, Edouard Grave, Armand Joulin, and Sainbayar
  Sukhbaatar.
\newblock Addressing some limitations of transformers with feedback memory.
\newblock \emph{arXiv preprint arXiv:2002.09402}, 2020.

\bibitem[Fedus et~al.(2021)Fedus, Zoph, and Shazeer]{fedus2021switch}
William Fedus, Barret Zoph, and Noam Shazeer.
\newblock Switch transformers: Scaling to trillion parameter models with simple
  and efficient sparsity.
\newblock \emph{arXiv preprint arXiv:2101.03961}, 2021.

\bibitem[Lewis et~al.(2021)Lewis, Bhosale, Dettmers, Goyal, and
  Zettlemoyer]{lewis2021base}
Mike Lewis, Shruti Bhosale, Tim Dettmers, Naman Goyal, and Luke Zettlemoyer.
\newblock Base layers: Simplifying training of large, sparse models.
\newblock \emph{arXiv preprint arXiv:2103.16716}, 2021.

\bibitem[Al-Rfou et~al.(2019)Al-Rfou, Choe, Constant, Guo, and
  Jones]{al2018character}
Rami Al-Rfou, Dokook Choe, Noah Constant, Mandy Guo, and Llion Jones.
\newblock Character-level language modeling with deeper self-attention.
\newblock In \emph{Proceedings of the 33rd {AAAI} Conference on Artificial
  Intelligence}, 2019.

\bibitem[Shaw et~al.(2018)Shaw, Uszkoreit, and Vaswani]{shaw2018self}
Peter Shaw, Jakob Uszkoreit, and Ashish Vaswani.
\newblock Self-attention with relative position representations.
\newblock In \emph{{NAACL-HLT} {(2)}}, 2018.

\bibitem[Mahoney(2011)]{mahoney2011large}
Matt Mahoney.
\newblock Large text compression benchmark.
\newblock \emph{URL: http://www. mattmahoney. net/text/text. html}, 2011.

\bibitem[Sukhbaatar et~al.(2019)Sukhbaatar, Grave, Bojanowski, and
  Joulin]{sukhbaatar2019adaptive}
Sainbayar Sukhbaatar, {\'E}douard Grave, Piotr Bojanowski, and Armand Joulin.
\newblock Adaptive attention span in transformers.
\newblock In \emph{Proceedings of the 57th Annual Meeting of the Association
  for Computational Linguistics}, pages 331--335, 2019.

\bibitem[Yang et~al.(2018)Yang, Yuan, Cer, Kong, Constant, Pilar, Ge, Sung,
  Strope, and Kurzweil]{reddit_use}
Yinfei Yang, Steve Yuan, Daniel Cer, Sheng-yi Kong, Noah Constant, Petr Pilar,
  Heming Ge, Yun-Hsuan Sung, Brian Strope, and Ray Kurzweil.
\newblock Learning semantic textual similarity from conversations.
\newblock In \emph{Proceedings of The Third Workshop on Representation Learning
  for {NLP}}, pages 164--174, Melbourne, Australia, July 2018. Association for
  Computational Linguistics.

\bibitem[Mazar{\'e} et~al.(2018)Mazar{\'e}, Humeau, Raison, and
  Bordes]{mazare2018trainingmillions}
Pierre-Emmanuel Mazar{\'e}, Samuel Humeau, Martin Raison, and Antoine Bordes.
\newblock Training millions of personalized dialogue agents.
\newblock In \emph{Proceedings of the 2018 Conference on Empirical Methods in
  Natural Language Processing}, pages 2775--2779, Brussels, Belgium,
  October-November 2018. Association for Computational Linguistics.

\bibitem[Keskar et~al.(2019)Keskar, McCann, Varshney, Xiong, and
  Socher]{keskar2019ctrl}
Nitish~Shirish Keskar, Bryan McCann, Lav~R Varshney, Caiming Xiong, and Richard
  Socher.
\newblock {CTRL}: A conditional transformer language model for controllable
  generation.
\newblock \emph{arXiv preprint arXiv:1909.05858}, 2019.

\bibitem[Shuster et~al.(2019)Shuster, Ju, Roller, Dinan, Boureau, and
  Weston]{shuster2019dialogue}
Kurt Shuster, Da~Ju, Stephen Roller, Emily Dinan, Y-Lan Boureau, and Jason
  Weston.
\newblock The dialogue dodecathlon: Open-domain knowledge and image grounded
  conversational agents, 2019.

\bibitem[Humeau et~al.(2019)Humeau, Shuster, Lachaux, and
  Weston]{humeau2019polyencoder}
Samuel Humeau, Kurt Shuster, Marie{-}Anne Lachaux, and Jason Weston.
\newblock Poly-encoders: Architectures and pre-training strategies for fast and
  accurate multi-sentence scoring.
\newblock In \emph{Proceedings of the International Conference on Learning
  Representations}, 2019.

\bibitem[Baumgartner et~al.(2020)Baumgartner, Zannettou, Keegan, Squire, and
  Blackburn]{baumgartner2020pushshift}
Jason Baumgartner, Savvas Zannettou, Brian Keegan, Megan Squire, and Jeremy
  Blackburn.
\newblock The pushshift reddit dataset.
\newblock \emph{arXiv preprint arXiv:2001.08435}, 2020.

\bibitem[Roller et~al.(2020)Roller, Dinan, Goyal, Ju, Williamson, Liu, Xu, Ott,
  Shuster, Smith, et~al.]{roller2020recipes}
Stephen Roller, Emily Dinan, Naman Goyal, Da~Ju, Mary Williamson, Yinhan Liu,
  Jing Xu, Myle Ott, Kurt Shuster, Eric~M Smith, et~al.
\newblock Recipes for building an open-domain chatbot.
\newblock \emph{arXiv preprint arXiv:2004.13637}, 2020.

\bibitem[Liu et~al.(2019)Liu, Ott, Goyal, Du, Joshi, Chen, Levy, Lewis,
  Zettlemoyer, and Stoyanov]{liu2019roberta}
Yinhan Liu, Myle Ott, Naman Goyal, Jingfei Du, Mandar Joshi, Danqi Chen, Omer
  Levy, Mike Lewis, Luke Zettlemoyer, and Veselin Stoyanov.
\newblock Roberta: A robustly optimized bert pretraining approach.
\newblock \emph{arXiv preprint arXiv:1907.11692}, 2019.

\bibitem[Conneau et~al.(2019)Conneau, Khandelwal, Goyal, Chaudhary, Wenzek,
  Guzm{\'a}n, Grave, Ott, Zettlemoyer, and Stoyanov]{conneau2019unsupervised}
Alexis Conneau, Kartikay Khandelwal, Naman Goyal, Vishrav Chaudhary, Guillaume
  Wenzek, Francisco Guzm{\'a}n, Edouard Grave, Myle Ott, Luke Zettlemoyer, and
  Veselin Stoyanov.
\newblock Unsupervised cross-lingual representation learning at scale.
\newblock \emph{arXiv preprint arXiv:1911.02116}, 2019.

\end{thebibliography}
\bibliographystyle{unsrtnat}

\clearpage 
\newpage 

\appendix
\section{Appendix}

\subsection{Task Setups} \label{app:tasksetup}
\label{sec:hyperparams}

\begin{table}[h!]
    \caption{Shared hyperparameters for all models, given for each task.}
    \centering
    \begin{tabular}{lcccc}
    \toprule
    Hyperparameter & Random Walk & Algorithm & Reddit/BASE & Enwik8 \\
    \midrule
    Layers              & 4     & 4     & 8     & 8 \\  
    Hidden size         & 256   & 256   & 512  & 512 \\  
    Head count          & 4     & 4     & 8     & 8 \\    
    Dropout rate        & 0.2   & 0.2   & 0.3   & 0.3 \\
    Embed. dropout      & -     & -     & 0.2    & 0.2 \\
    BPTT (i.e. segment) len            & 128   & 128   & 256   & 256 \\
    Batch size          & 512   & 256   & 512  & 512 \\
    Learning rate (LR)  & 1e-4  & 1e-4  & 7e-4 & 7e-4 \\
    Gradient clip       & 0.1   & 0.1   & 0.1   & 0.1 \\
    LR warm-up steps    & 1k    & 1k    & 8k    & 8k \\
    \bottomrule
    \end{tabular}
    \label{app:hyper_lm}
\end{table}

We provide the hyperparameter setups shared across our models for each task  in \autoref{app:hyper_lm}. In addition, the hyperparameters tuned for each model for the best performance are shown in \autoref{app:best_params}, which were selected using validation performance. We also provide a textual description of some aspects of the base models below. 

\paragraph{Random Walk}
 We train 4-layer models with a hidden size of 256 and 4 attention heads. We use a learning rate of 1e-4 and 1000 warmup updates to train the models. They are trained for 50k updates with batch size 512. The global staircase models are trained for 400k updates since they need longer to converge.  We ran each setting 10 times, except for the Cached Staircase model which was run 5 times.

\paragraph{Algorithm}
 We train the 4-layer model with a hidden size of 256 and 4 attention heads. Models are trained to 100k updates with batch size of 256 and learning rate of 1e-4, 1000 warmup updates. We train the global staircase models for 400k steps. We ran each setting 10 times, except for the Cached Staircase model which was run 5 times.

\paragraph{Pushshift.io Reddit}
 We train 8-layer models with hidden size of 1024, 8 attention heads. They are trained for 100k updates with a learning rate of 7e-4, 8000 warmup updates and a batch size of 512.

\paragraph{BASE Data}
 We train 8-layer models with hidden size of 1024, 8 attention heads. They are trained for 80k updates with a learning rate of 7e-4, 8000 warmup updates and a batch size of 512.

\paragraph{Enwik8}
 We train 8-layer models with 8 attention heads. They are trained for 100k updates with a learning rate of 7e-4, 8000 warmup updates and a batch size of 512.

\begin{table}[t]
   \caption{Hyperparameters for best performing models across all tasks. }
    \centering
    \begin{tabular}{llccccc}
    \toprule
    Tasks & Models & Recurrent & Step & Forward & Attention\\
    & & steps  & size & size & span ($S$)  \\
    \midrule
     & Staircase & 8 & 64 & 8 & -\\
    Random Walk & Cached staircase & 1 & 256 & 4 & - \\
     & Ladder & 8 & - & - & 512\\
    \midrule
     & Staircase & 8 & 64 & 8 & -\\
    Algorithm & Cached staircase & 1 & 64 & 4 & -\\
     & Ladder & 2 & - & - & 512\\
    \midrule
     & Staircase & 8 & 256 & 32 & - \\
    Reddit & Cached staircase & 1 & 384 & 128 & -\\
     & Ladder & 8 & - & - & 256\\
    \midrule
     & Staircase        & 8 & 256 & 32 & - \\
    BASE Data  & Cached staircase & 1 & 384 & 128 & -\\
     & Ladder           & 8 & - & - & 256\\
    \midrule
     & Staircase & 4 & 256 & 64 & - \\
    Enwik8 & Cached staircase & 1 & 260 & 4 & -\\
     & Ladder & 4 & - & - & 256\\
    \bottomrule
    \end{tabular}
    \label{app:best_params}
\end{table}

\subsection{Further Detailed Results}

Detailed results for a number of our tasks beyond those results reported in the main paper are provided in 
Tables \ref{tab:algorithm-performance},  \ref{tab:random_walk}, 
\ref{episodic-reddit} and \ref{enwiki}.

\begin{table}[t]
  \caption{Algorithm task detailed results.}
  \label{tab:algorithm-performance}
  \centering
  \begin{tabular}{llllrr}
    \toprule
    Models & Recurrent & Step & Forward & Valid  & Test  \\
         & steps  & size & size & (err. \%) & (err. \%) \\
    \midrule
Transformer-XL & - & - & - & 59.1 $\pm$ 12.5 & 59.1 $\pm$ 12.4 \\
Feedback Trans. & 1 & 32 & 1 & 0.3 $\pm$ 0.0 & 0.3 $\pm$ 0.0 \\ \midrule
Staircase & 8 & 64 & 8 & 12.8 $\pm$ 16.5 & 12.6 $\pm$ 16.2 \\
Staircase & 4 & 64 & 16 & 0.5 $\pm$ 0.5 & 0.5 $\pm$ 0.7 \\
Staircase & 2  & 64 & 32 & 0.6 $\pm$ 0.2 & 0.5 $\pm$ 0.2 \\
\midrule
Cached Staircase & 1 & 64 & 4 & 24.6 $\pm$ 21.8 & 24.3 $\pm$ 21.7 \\
Cached Staircase & 1 & 64 & 8 & 31.7 $\pm$ 28.7 & 31.3 $\pm$ 28.7\\
Cached Staircase & 1 & 64 & 16 & 27.8 $\pm$ 13.6 & 27.3 $\pm$ 13.6 \\
    \midrule
Global Cached Staircase & 1 & 512 &  8 & 0.0 $\pm$ 0.1 & 0.0 $\pm$ 0.1 \\
Global Cached Staircase & 1 & 512 & 16 & 7.1 $\pm$ 19.3 & 7.1 $\pm$ 19.3\\
Global Cached Staircase & 1 & 512 & 32 & 20.0 $\pm$ 23.0 & 19.7 $\pm$ 22.7 \\
    \midrule
Ladder & 2 & - & - & 64.6 $\pm$ 15.5 & 64.4 $\pm$ 15.3 \\
Ladder & 4 & - & - & 74.3 $\pm$ 9.7 & 74.0 $\pm$ 9.4 \\
Ladder & 8 & - & - & 78.9 $\pm$ 6.4 & 78.6 $\pm$ 6.5 \\
    \bottomrule
  \end{tabular}
\end{table}
\begin{table}[h]
  \caption{Random Walk task detailed results.}
  \label{tab:random_walk}
  \centering
  \begin{tabular}{llllrr}
    \toprule
        Models & Recurrent & Step & Forward & Valid  & Test  \\
         & steps  & size & size & (\%) & (\%) \\
    \midrule
Transformer-XL & 1 & - & - & 90.1 $\pm$ 4.6 & 90.1 $\pm$ 4.6 \\ 
Feedback Trans. & 1 & 64 & 1 & 0.1 $\pm$ 0.0 & 0.1 $\pm$ 0.0 \\
\midrule
Staircase & 8 & 64 & 8 & 0.2 $\pm$ 0.1 & 0.2 $\pm$ 0.1 \\
Staircase & 4 & 64 & 16 & 0.2 $\pm$ 0.2 & 0.2 $\pm$ 0.2 \\
Staircase & 2 & 64 & 32 & 1.0 $\pm$ 1.3 & 1.0 $\pm$ 1.2 \\
\midrule
Cached Staircase & 1 & 256 & 4 & 0.1 $\pm$ 0.0 & 0.1 $\pm$ 0.0\\
Cached Staircase & 1 & 256 & 8 & 1.9 $\pm$ 2.0 & 1.9 $\pm$ 2.0 \\
Cached Staircase & 1 & 256 & 16 & 27.2 $\pm$ 8.0 & 27.3 $\pm$ 8.2 \\
    \midrule
Global Cached Staircase & 1 & 512 & 8 & 0.0 $\pm$ 0.0 & 0.0 $\pm$ 0.0 \\
Global Cached Staircase & 1 & 512 & 16 & 1.4 $\pm$ 0.6 & 1.3 $\pm$ 0.5\\
Global Cached Staircase & 1 & 512 & 32 & 52.4 $\pm$ 16.4 & 52.4 $\pm$ 16.4 \\
    \midrule
Ladder & 2 & - & - & 75.2 $\pm$ 12.5 & 75.1 $\pm$ 12.6 \\
Ladder & 4 & - & - & 35.6 $\pm$ 5.3 & 35.5 $\pm$ 5.3 \\
Ladder & 8 & - & - & 3.1 $\pm$ 1.6 & 3.2 $\pm$ 1.6 \\
    \bottomrule
  \end{tabular}
\end{table}

\begin{table}[t]
  \caption{Results on pushshift.io Reddit with Episodic data. Here,  we perform experiments where we prepare an episodic version of the data, where we keep the text length fixed to 256 BPE tokens. The shorter episodes are padded, and longer ones are split into two.}
  \label{episodic-reddit}
  \centering
  \begin{tabular}{lccccc}
    \toprule
    Model & Recurrent & Step & Forward & Valid. & Test \\
     & steps  & size & size & (ppl) & (ppl) \\
    \midrule
    Transformer-XL         & - & - & 256 & 27.6 & 27.3\\
    Cached Staircase & 1 & 256 & 32 & 27.9 & 27.6 \\
    Cached Staircase & 1 & 256 & 64 & 27.8 & 27.6 \\
    Cached Staircase & 1 & 256 & 128 & 27.6 & 27.3 \\
    Staircase & 2 & 256 & 128 & 26.7 & 26.4 \\
    Staircase & 4 & 256 & 64 & 25.2 & 24.9 \\
    Staircase & 8 & 256 & 32 & 24.3 & 24.0 \\
    \bottomrule
  \end{tabular}
\end{table}

\begin{table}[t]
  \caption{Enwik8 task detailed results.} 
  \label{enwiki}
  \centering
  \begin{tabular}{llllll}
    \toprule
     Models & Recurrent  & Step & Forward & Valid & Test \\
     &steps & size & size & (ppl) & (ppl) \\
    \midrule
Transformer-XL & -  & 256 & 256 & 1.17 & 1.15 \\ 
Feedback Trans. & 1 & 256 & 1 & 1.13 & 1.12 \\ 
Cached Staircase & 1  & 260 & 4 & 1.14 & 1.13\\
Cached Staircase & 1 & 288 & 32 & 1.15 & 1.13\\
Cached Staircase & 1 & 320 & 64 & 1.15 & 1.13\\
Cached Staircase & 1 & 384 & 128 & 1.15 & 1.13 \\
\midrule
Staircase & 2 & 256 & 128 & 1.15 & 1.14 \\
Ladder & 2 & - & - & 1.14 & 1.12 \\
\midrule
Staircase & 4 & 256 & 64 & 1.14 & 1.14 \\
Ladder & 4 & - & - & 1.12 & 1.11 \\
    \bottomrule
  \end{tabular}
\end{table}

\section{Computational Resources}
\label{sec:compute}

All experiments were run in an internal cluster using 32GB V100 GPUs.The usage varies on recurrent steps; here, we list the maximum resources used in experiments (e.g. 8x ladder)

\begin{itemize}
    \item Random Walk experiment used maximum 8 GPUs for $\sim$7 hours. %We ran each setting 10 times, except for the Cached Staircase model which was run 5 times.
    \item Algorithm experiment used maximum 2 GPUs for $\sim$30 hours. %We ran each setting 10 times, except for the Cached Staircase model which was run 5 times.
    \item Language modeling experiments used maximum 32 GPUs for $\sim$30 hours. Experiments were run only once.
\end{itemize}

\section{Limitations, Scope and Societal Impact}
\label{sec:socimpact}

Improvements to language modeling could have implications on a large number of surfaces across humanity. Additionally, our experiments show how to obtain higher performance by using more compute (with the same number of parameters), therefore increasing carbon emissions. The datasets used contain PII and offensive content in the form of text, as they were originally procured from the internet (we do not release any data in this work).

Our family of models improves perplexity in language modeling and tracking of state in certain tasks compared to methods with  the same number of model parameters, but typically costs more in terms of compute. 
Therefore, whether this is beneficial depends on the end application and user requirements. E.g., these methods are good in memory-bound setups, but  when compute-bound it will depend on the precise setup.

\end{document}